\documentclass[lettersize,journal]{IEEEtran}
\usepackage{amsmath,amsfonts}
\usepackage{algorithmic}
\usepackage{algorithm}
\usepackage{array}
\usepackage[caption=false,font=normalsize,labelfont=sf,textfont=sf]{subfig}
\usepackage{textcomp}
\usepackage{stfloats}
\usepackage{url}
\usepackage{verbatim}
\usepackage{graphicx}
\usepackage{cite}
\usepackage{hyperref}
\hypersetup{hidelinks}
\usepackage{bm} 
\def\BibTeX{{\rm B\kern-.05em{\sc i\kern-.025em b}\kern-.08em
    T\kern-.1667em\lower.7ex\hbox{E}\kern-.125emX}}

\begin{document}

\title{PSG-MAE: Robust Multitask Sleep Event Monitoring using Multichannel PSG Reconstruction and Inter-channel Contrastive Learning}

\author{Yifei Wang, Qi Liu, \IEEEmembership{Senior Member, IEEE}, Fuli Min, and Honghao Wang
\thanks{This work was supported in part by the National Natural Science Foundation of China under Grant 62202174, in part by the Basic and Applied Basic Research Foundation of Guangzhou under Grant 2023A04J1674, in part by The Taihu Lake Innovation Fund for the School of Future Technology of South China University of Technology under Grant 2024B105611004, and in part by Guangdong Science and Technology Department Grant 2024A1313010012. \textit{(Corresponding author: Qi Liu and Honghao Wang.)}}
\thanks{Yifei Wang and Qi Liu are with the School of Future Technology, South China University of Technology, Guangzhou 511400, China (e-mail: ywang634@outlook.com; drliuqi@scut.edu.cn)}
\thanks{Fuli Min and Honghao Wang are with the Department of Neurology, Guangzhou First People's Hospital, and School of Medicine, South China University of Technology, Guangzhou 510180, China (e-mail: minfuli@163.com; wang\_whh@163.com)}}

\maketitle

\begin{abstract}
Polysomnography (PSG) signals are essential for studying sleep processes and diagnosing sleep disorders. Analyzing PSG data through deep neural networks (DNNs) for automated sleep monitoring has become increasingly feasible. However, the limited availability of datasets for certain sleep events often leads to DNNs focusing on a single task with a single-sourced training dataset. As a result, these models struggle to transfer to new sleep events and lack robustness when applied to new datasets. To address these challenges, we propose PSG-MAE, a mask autoencoder (MAE) based pre-training framework. By performing self-supervised learning on a large volume of unlabeled PSG data, PSG-MAE develops a robust feature extraction network that can be broadly applied to various sleep event monitoring tasks. Unlike conventional MAEs, PSG-MAE generates complementary masks across PSG channels, integrates a multichannel signal reconstruction method, and employs a self-supervised inter-channel contrastive learning (ICCL) strategy. This approach enables the encoder to capture temporal features from each channel while simultaneously learning latent relationships between channels, thereby enhancing the utilization of multichannel information. Experimental results show that PSG-MAE effectively captures both temporal details and inter-channel information from PSG signals. When the encoder pre-trained through PSG-MAE is fine-tuned with downstream feature decomposition networks, it achieves an accuracy of 83.7\% for sleep staging and 90.45\% for detecting obstructive sleep apnea, which highlights the framework's robustness and broad applicability.
\end{abstract}


\section{Introduction}
\label{sec:introduction}
\IEEEPARstart{S}{leep} is an essential necessity for life maintenance. Consistent and adequate rest is crucial for improving health, productivity, well-being and quality of life as well as public safety \cite{ramar2021sleep}. In recent years, the accelerated pace of global urbanization and rising stress have exacerbated the prevalence of sleep disorders, posing a substantial challenge to public health. Common sleep disorders, such as insomnia, arousal disorders, sleep apnea, rapid eye movement sleep behavior disorder (RBD), and periodic limb movement disorder (PLMD), are associated with a heightened risk of medical complications, including cardiovascular diseases, depression, and anxiety, diabetes and compromised immune function \cite{lee2024automatic}. Consequently, the development of automated screening and intervention methods for sleep disorders is of significant research value.

\begin{figure}[!t]
\centerline{\includegraphics[width=0.9\columnwidth]{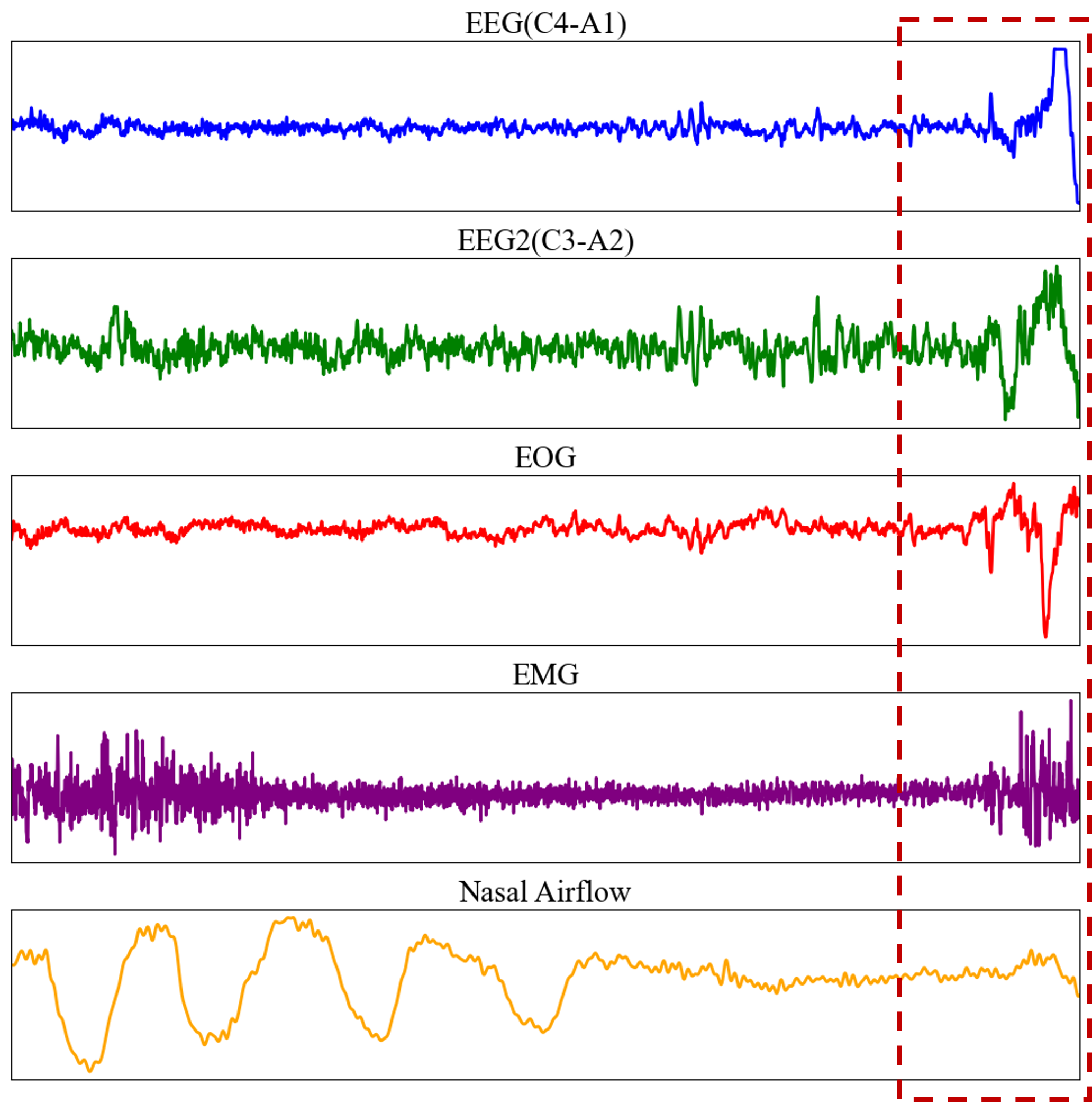}} 
\caption{Polysomnography (PSG) of one sleep epoch (30s), during which arousal occurs, marked in the dashed box. A sleep event during sleep often induces abrupt fluctuations in multiple channels of the PSG signals. Integrating the variations across different channels can help improve the accuracy of sleep event monitoring.}
\label{fig 1}
\end{figure}

Currently, combining polysomnography (PSG) with deep neural networks (DNNs) has become a widely explored approach in automated sleep monitoring research \cite{yazdi2024review}. PSG is widely considered the most reliable method in the field of sleep medicine for diagnosing sleep-related disorders, often employed to evaluate both the diagnosis and efficacy of treatment for sleep disturbances \cite{RUNDO2019381}, \cite{zhou2024interpretable}. A standard PSG recording gathers data on brain waves (electroencephalography, EEG), eye movements (electrooculography, EOG), chin and leg muscle activity (electromyography, EMG), heart activity (electrocardiography, ECG), chest and abdominal breathing effort, nasal airflow, oxygen saturation, etc \cite{zhang2022auto}. While polysomnography (PSG) provides comprehensive documentation of sleep patterns, the analysis and clinical interpretation of these neurophysiological recordings demand rigorous systematic training. Additionally, annotating PSG data is labor-intensive and time-consuming, with 2-3 experts typically spending about 2 hours to annotate an 8-hour sleep recording. Subjective differences among experts can also lead to variability in annotation results \cite{perez2020future}, \cite{khalili2021automatic}. The process of annotating PSG sleep data includes the categorization of sleep stages and the identification of sleep events. In accordance with the sleep staging guidelines outlined by the American Academy of Sleep Medicine (AASM), PSG data is divided into 30-second segments along the temporal dimension. Each epoch is then classified into stages, including wake (W), non-rapid eye movement (NREM) stages (N1, N2, N3), and rapid eye movement (REM) sleep \cite{berry2015aasm}. The epoch-based segmentation approach is utilized in contemporary clinical practice to label sleep events, including sleep apnea and limb movements, in order to ensure consistent analysis across research studies.

The multichannel nature of PSG signals makes them well-suited for integration with machine learning and deep neural networks, enabling the automatic extraction of complex sleep features and effectively modeling nonlinear relationships for accurate sleep event annotation \cite{somanna2019automated}, \cite{de2024sleep}. Current PSG-driven automated sleep events monitoring can be divided into two main areas. One focuses on the automatic sleep staging \cite{satapathy2023machine},\cite{prochazka2017adaptive},\cite{zhang2024swinsleep}, while the other involves the detection and labeling of sleep behaviors, events, and disorders \cite{li2018feature},\cite{bartolo2001arrhythmia}. However, there are two main challenges in current sleep event monitoring models. First, the wide variety of sleep events and their different manifestations across populations result in a limited quantity of public datasets for certain sleep events \cite{ehrlich2024state},\cite{lee2022large}. As a result, many models are trained on small, task-specific datasets, making them sensitive to the feature distribution of the data. This limits their ability to transfer to other sleep event monitoring tasks and hinders a comprehensive, multidimensional evaluation of sleep. Second, most current sleep models rely on only a single or few PSG signal channels for specific monitoring tasks, neglecting the potential inter-channel interactions. As shown in Fig. \ref{fig 1}, sleep events often induce signal changes across multiple channels. By integrating information from multiple channels, we can reduce misjudgments caused by disturbances in individual channels while improving the overall accuracy of event detection.

To address the challenges mentioned above, we propose leveraging unlabeled data through self-supervised learning to improve the model's stability in PSG feature extraction and enhance its performance in multichannel information fusion. To this end, we introduce PSG-MAE, a novel pre-training framework for PSG signals. In contrast to normal MAEs, PSG-MAE is based on a complementary-masking strategy for multichannel PSG signal reconstruction, meaning that one epoch (30 seconds) of PSG data is used as input with a pair of complementary masks generated along the channel dimension. Based on this design, we not only present a redesigned channel-level reconstruction loss but also introduce inter-channel contrastive learning (ICCL) to further explore the inter-channel interaction information. PSG-MAE aims not only to capture fine-grained temporal information but also to learn the potential relationships between multiple channels of PSG. After the pre-training phase of PSG-MAE, a robust PSG encoder is built, which can be combined with downstream feature decomposition networks and fine-tuned to adapt to different sleep event monitoring tasks. The contributions of this paper can be summarized as follows:
\begin{itemize}
\item We propose PSG-MAE, a novel pre-training framework for PSG signals, which employs a complementary-masking strategy and leverages unlabeled PSG data for self-supervised learning. This approach enhances the feature extraction process, which is applicable to a wide range of sleep event monitoring tasks.

\item To better exploit the multichannel nature of PSG signals, we introduce an updated channel-level signal reconstruction loss and a novel ICCL method. These innovations improve PSG-MAE's ability to capture fine-grained temporal information from each channel while also effectively modeling inter-channel interactions.

\item The pre-trained PSG encoder demonstrates exceptional discriminative performance and robustness across multiple downstream sleep event monitoring tasks, including sleep staging and obstructive sleep apnea (OSA) detection, showcasing its broader applicability compared to traditional single-task models.

\end{itemize}

\begin{figure*}[!t]
\centering
\includegraphics[width=\textwidth]{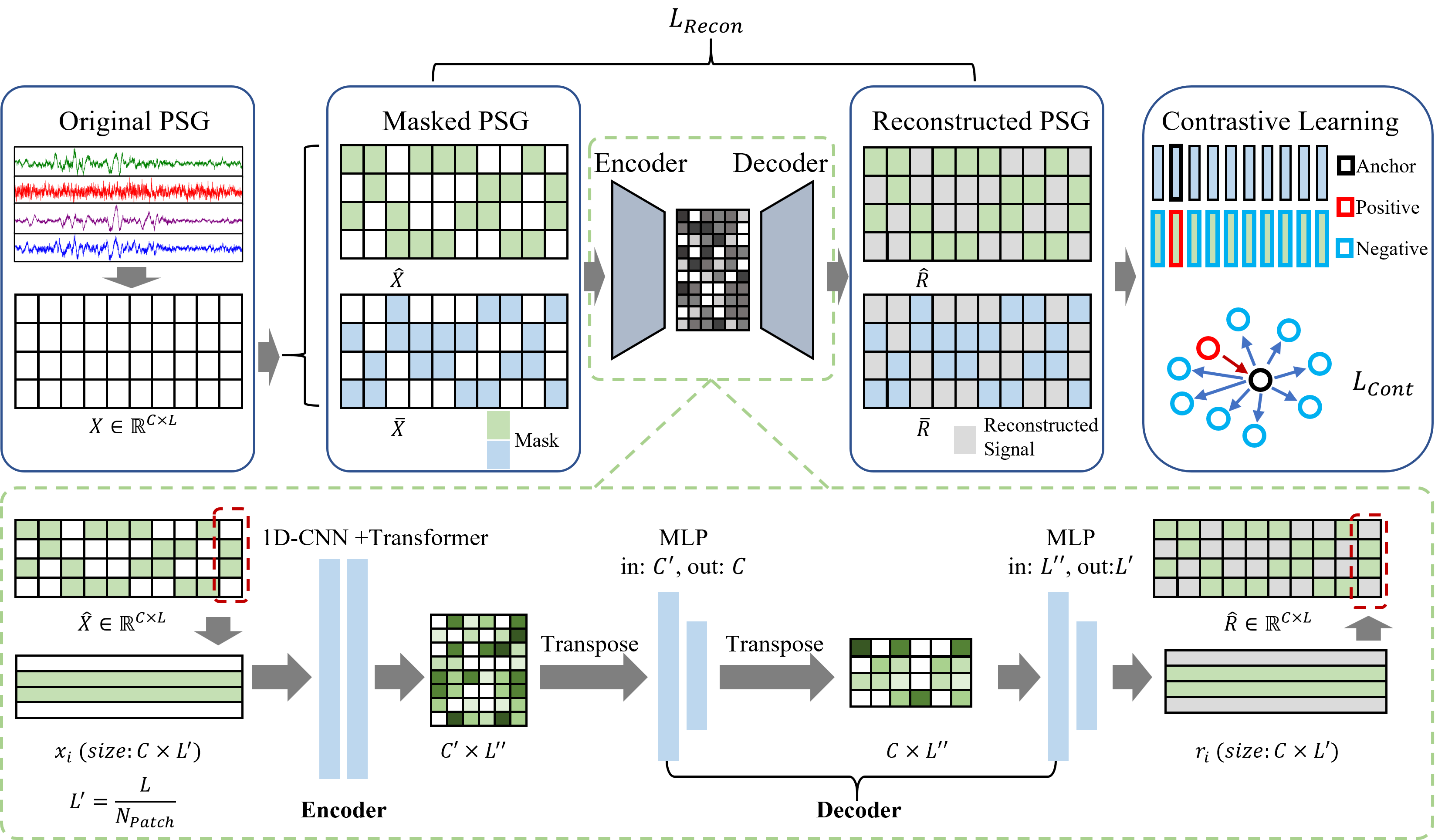}
\caption{The framework of \textbf{PSG-MAE}: The original PSG signal is divided into subsegments along the time dimension, followed by the application of complementary masks across the channel dimension. After passing through the encoder-decoder network, the unmasked portions of the signal are reconstructed, with the channel-level reconstruction loss facilitating the learning of temporal features in the original signal. In the pair of reconstructed PSG signals, one sub-segment is treated as an anchor, whose corresponding sub-segment in the other signal is considered as a positive sample, while the remaining subsegments are negative samples. ICCL is then applied to learn the intrinsic relationships between different channels by maximizing the distance of positive pairs and minimizing that of negative ones.}

\label{fig2}
\end{figure*}

\section{Related Work}

The research on PSG-driven automated sleep event monitoring currently focuses primarily on addressing the issues of sleep staging and the identification of other sleep events. Research on automated sleep staging generally follows two approaches: using raw signals as input and applying time-frequency transformations (e.g., Fourier transform, continuous wavelet transform) to generate spectrum as input\cite{sekkal2022automatic}. Each channel of PSG signals is typically represented as one-dimensional time-series data. Consequently, feature extraction modules commonly employ models such as 1D convolutional neural networks (1D-CNNs) and recurrent neural networks (RNNs), which are well-suited for processing one-dimensional features and capturing temporal information. N. Goshtasbi et al. introduced SleepFCN, a fully convolutional framework that utilizes dual 1D-CNN branches with distinct kernel sizes to capture information across various EEG frequency bands \cite{goshtasbi2022sleepfcn}. Building upon SleepFCN, H. Zhu et al. developed MS-HNN, which integrates a squeeze-and-excitation (SE) block into the dual-kernel 1D-CNN branches to select more informative features. Additionally, MS-HNN employs bidirectional gated recurrent units (Bi-GRU) in the downstream network to learn temporal dependencies \cite{zhu2023ms}. Y. Na et al. proposed using convolutional layers to fuse multichannel PSG data, allowing various physiological signals to contribute to the decision-making process \cite{na2022evaluation}. Physiological signals, such as EEG, exhibit distinct variations in frequency bands and power distributions across different sleep stages. Time-frequency transformations effectively capture these frequency characteristics, especially for applications involving non-stationary signals. P. Huy et al. introduced SeqSleepNet, which uses short-time Fourier transform (STFT) to convert PSG signals into time-frequency spectrograms and applies recurrent layers to capture both short-term and long-term dependencies within each epoch \cite{huy2019seqsleepnet}. Y. Dai et al. proposed generating multichannel time-frequency spectrograms and employing multiple transformer groups to capture both individual channel features and joint features across channels \cite{dai2023multichannelsleepnet}.

Research on automated monitoring of other sleep events, such as sleep disorders, has advanced significantly. X. Zhao et al. segmented signals from the C3-A2 and C4-A1 EEG channels into five sub-bands, extracted entropy and variance features, and used machine learning to classify obstructive sleep apnea (OSA), central sleep apnea (CSA), and normal breathing events with \cite{zhao2021classification}. A. Brink-Kjaer et al. used CNN+Bi-LSTM to extract features and temporal information from 5-minute PSG epochs for RBD classification, extending it with latent space transfer to analyze entire night recordings \cite{brink2022end}. W. Qu et al. combined single-channel EEG data with a domain adaptation strategy, using similarity loss between encoders from source and target domains to learn temporal features. The source encoder was then integrated with LSTM networks for insomnia detection. \cite{qu2021single}. 

Self-supervised learning has been shown to improve the robustness of feature extraction by leveraging unlabeled data. In this context, MAE learns robust feature representations by masking and reconstructing portions of the input signals, and has demonstrated superior performance in various downstream tasks \cite{he2022masked},\cite{wang2023videomae},\cite{cai2023marlin},\cite{zhang2024mart}. MAE has been applied to the representation learning of temporal physiological signals. Y.-T. Lan et al. proposed the Corrupted Emotion Autoencoder (CEMOAE) framework to address channel corruption in EEG topographic maps by reconstructing masked signals to learn robust features and fine-tuning a pre-trained autoencoder for emotion recognition \cite{lan2024cemoae}. H. Ma et al. proposed a novel Region-State Masked Autoencoder (RS-MAE) that reduces redundancy in dynamic functional connectivity matrices, introduces region-state embeddings, and applies data augmentation to enhance classification performance for neuropsychiatric disorders based on resting-state fMRI. The encoder, pre-trained in this manner, has been shown to improve downstream task performance by capturing more relevant features \cite{ma2024rs}.

\section{Methods}
\subsection{Overview}

The general framework of PSG-MAE, based on the complementary masking strategy, is shown in Fig. \ref{fig2}. A 30-second multichannel PSG data segment, divided into temporally equal-length subsegments, serves as the input. After that, the input is processed by applying a pair of randomly generated and complementary masks, forming two masked inputs that are fed into a shared encoder-decoder network to reconstruct the unmasked regions. The model then applies multichannel reconstruction and self-supervised ICCL to capture both temporal features and interactions among channels within the PSG input.

\subsection{Multichannel Signal Reconstruction with Complementary-masking}
The input PSG data \(X \in \mathbb{R}^{C \times L}\) has \(C\) channels, with each channel containing \(L\) time steps. According to the standards of the International Classification of Sleep Disorders (ICSD), \(X\) encompasses a 30-second window of PSG data, where \(L = 30s \times \text{sampling frequency}\). The input \(X\) is partitioned into smaller and manageable subsegments. These subsegments are defined by a hyperparameter \(N_{Patch}\), which specifies how many patches the original time series data should be divided into along the time dimension. resulting in a set of subsegments \(x_i\). This process can be expressed as

\begin{equation}
X = [x_1, x_2, x_3, \dots, x_N], \ x_i \in \mathbb{R}^{C \times L'}, \ L' = \frac{L}{N_{Patch}}.
\end{equation}
At the beginning of the pre-training phase, the PSG-MAE framework generates a pair of complementary masks, \( M \) and \( (\mathbf{1} - M) \in \mathbb{R}^{C \times L} \), each having the same size as the original input. The masking process begins by randomly selecting the floor of half the number of channels (\( \lfloor C/2 \rfloor \)) from each sub-segment \( x_i \), and then combining all selected channels to form the mask \( M \), while the channels that are not selected form the complementary mask \( (\mathbf{1} - M) \). In this way, the two masks are complementary across the channel positions. Once the two complementary masks are created, the input matrix \( X \) is masked by applying both \( M \) and \( (\mathbf{1} - M) \) across all subsegments, a pair of inputs to the shared encoder is generated as

\begin{equation}
\hat{X} = M \ast X,
\end{equation}
\begin{equation}
\bar{X}=(\mathbf{1}-M)\ast X.
\end{equation}
The masked data \(\hat{X}\) and \(\bar{X}\), after undergoing the masking process, are passed through a transformer-based encoder for feature extraction and sequence modeling. This encoder leverages the self-attention mechanism to capture long-range dependencies within the data, enabling it to understand complex temporal patterns across multiple channels. After encoding, the transformed representation is fed into a multilayer perceptron (MLP)-based decoder. The decoder reconstructs the original signals from the encoded feature maps, producing a pair of reconstructed signals\(\hat{R}\) and \(\bar{R}\). These reconstructed signals are subsequently compared with the original signals to compute the redesigned channel-level reconstruction loss, formulated as 
\begin{equation}
L_{Recon}=L_{COS}+L_{MSE},
\label{Lrecon}
\end{equation}
\begin{equation}
L_{COS}=\frac{1}{C}\sum_{c=1}^{C}L_{{channelCOS_c}},
\label{cos_1}
\end{equation}
\begin{equation}
L_{channelCOS_c}=1-\frac{1}{N}\sum_{n=1}^{N}CosineSimilarity\left(r_n^c,x_n^c\right),
\label{cos_2}
\end{equation}
\begin{equation}
CosineSimilarity\left(r_n^c,x_n^c\right)=\frac{\sum_{t=1}^{N_{patch}}r_n^c\left(t\right)x_n^c\left(t\right)}{\|r_n^c \|\|x_n^c \|},
\label{cos_3}
\end{equation}
\begin{equation}
L_{MSE}=\frac{1}{C}\sum_{c=1}^{C}{\frac{1}{T}\sum_{t=1}^{T}\left(x^c\left(t\right)-r^c\left(t\right)\right)^2},
\label{lmse}
\end{equation}
where reconstruction loss is formulated as a combination of channel-level cosine similarity loss \(L_{COS}\) and channel-level mean squared error (MSE) loss \(L_{MSE}\). The \(L_{COS}\) is computed by firstly evaluating the cosine similarity of the \(c\)-th channel between the reconstructed signal subsegment \(r_n^c\) and the corresponding channel of the masked signal subsegment \(x_n^c\), n means the \(n\)-th segment, as expressed in \eqref{cos_3}. The resulting cosine similarities are then averaged across both the subsegments and channels, as outlined in \eqref{cos_2} and \eqref{cos_1}. The \(L_{MSE}\) is computed by first calculating the squared difference between the corresponding values of the masked signal  \(x^c(t)\) and the reconstructed signal \(r^c(t)\) for each time step \(t\) along the channel \(c\). The squared differences are then averaged over all \(T\) time steps and further averaged across all \(C\) channels, shown in \eqref{lmse}. By integrating these two channel-level loss functions, the cosine similarity loss enforces the preservation of the overall pattern and trend of the reconstructed signals relative to the original data, while the MSE loss refines the relative magnitudes of the numerical values. This synergy between the two losses ensures that the model captures both the structural integrity and the numerical accuracy of the signals, leading to a faithful reconstruction of the original data.

\subsection{Inter-channel Contrastive Learning}

During sleep monitoring, the physiological data recorded by PSG is multi-source, encompassing signals collected from various sensors (such as EEG, EOG, respiratory airflow sensors, EMG, etc.) distributed across the body. These signal channels are typically synchronized in the time domain so that sleep events are often reflected simultaneously across multiple signal channels. This temporal alignment enables the multi-dimensional data to exhibit complementary information characteristics in multi-task sleep event detection. To fully exploit this synergy, PSG-MAE introduces a novel inter-channel contrastive learning (ICCL) strategy, which allows the encoder to uncover the latent commonalities and differences between signals, thereby enhancing the collaborative representation of features across different channels. 

Specifically, we apply contrastive learning to the two groups of reconstructed output subsegments that form \(\hat{R}\) and \(\bar{R}\). The objective is to ensure that signal blocks from different channels, which correspond to the same time frame, (i.e., originate from the same subsegment), are drawn closer together in the feature space. In contrast, channel blocks from different time frames, which have weak correlations, are pushed further apart. Upon obtaining the reconstructed signals \(\hat{R}\) and \(\bar{R}\) from the shared decoder, we recursively select subsegment \( \hat{r}_i \) from\(\hat{R}\) as anchor sample.Then the corresponding subsegment \( \bar{r}_i \) from \(\bar{R}\), which contains complementary channel information relative to \( \hat{r}_i \) within the same time interval, is designated as the positive sample. Meanwhile, the remaining subsegments \( \hat{r}_{j \neq i} \) from \(\hat{R}\) serve as negative samples. During training, a triplet loss is employed to measure the relative distances among the anchor, positive, and negative samples. This strategy enables PSG-MAE to effectively learn and extract shared features across different channels while maintaining the independence and distinctiveness of temporal information. The channel contrastive loss \( L_{\text{CL}} \) is defined as
\begin{equation}
L_{\text{CL}} = \frac{1}{N_{patch}} \sum_{i=0}^{N_{patch}} F_{MAX},
\end{equation}

\begin{equation}
F_{MAX} = \max \left( 0, d(\hat{r}_i, \bar{r}_i) - \frac{1}{N_{patch}-1} \sum_{i \neq j} d(\hat{r}_i, \hat{r}_j) + \alpha \right),
\label{fmax}
\end{equation}

\begin{equation}
d(x, y) = \sqrt{\sum_{k=1}^{D} (x_k - y_k)^2},
\end{equation}
where \( d(x, y) \) denotes the Euclidean distance between samples \( x \) and \( y \).  \( x \) and \( y \) are two sample vectors with \( D \) dimensions. The components \( x_k \) and \( y_k \) represent the values of the samples in the \( k \)-th dimension. in \eqref{fmax}, \( d(x, y) \) measures the similarity between pairs of signal blocks, and \( \alpha \) is a hyperparameter that defines the minimum margin between the positive and negative samples, ensuring that the negative samples are sufficiently far away from the anchor in the feature space.

The loss function for the pre-training framework for PSG signals based on a complementary-masking strategy for multichannel signal reconstruction is defined as  
\begin{equation}
L = L_{\text{Recon}} + L_{\text{CL}}.
\end{equation}
By jointly optimizing these two losses, PSG-MAE effectively preserves the fine-grained temporal details of PSG signals while also capturing the correlated features across different channels. The resulting encoded features provide a robust intermediate representation that can be leveraged for a wide array of downstream sleep-related tasks, including sleep stage classification, OSA detection, and other sleep events recognition. This enables the encoder to serve as a flexible and adaptable component that can be integrated into various temporal feature decomposition networks, each tailored to meet the specific requirements of sleep monitoring and classification. Consequently, the pre-trained encoder can be further fine-tuned for different sleep analysis applications, thereby enhancing model performance across diverse datasets and task-specific scenarios.

\subsection{Downstream multitask sleep events monitoring}

\begin{figure}[!h]
\centerline{\includegraphics[width=0.7\columnwidth]{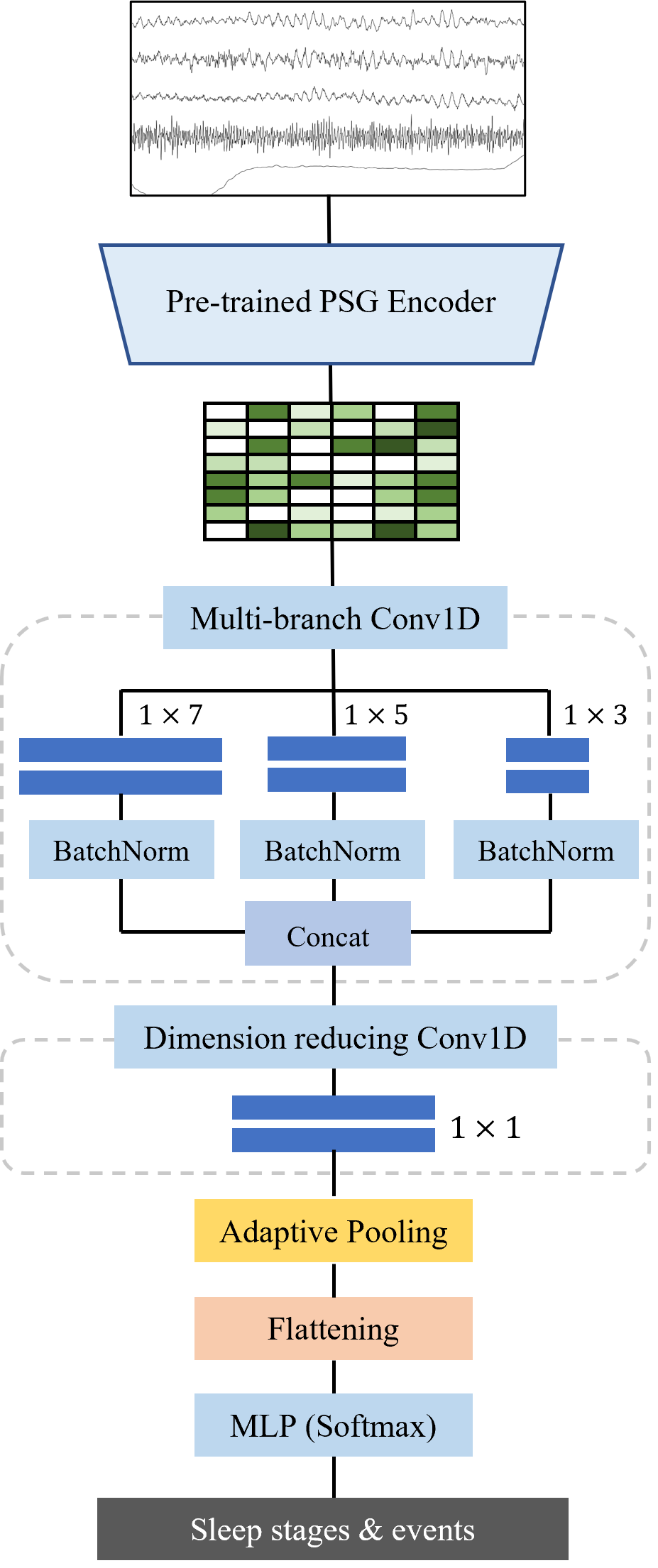}}
\caption{Basic structure of downstream sleep events monitoring network.}
\label{fig3}
\end{figure}

To effectively apply the pre-trained PSG-MAE encoder for downstream sleep event monitoring tasks, we design a feature decomposing network as illustrated in Fig. \ref{fig3}, this downstream network comprises several key components aimed at extracting and refining task-relevant information. To maximize the utility of the pre-trained features derived from PSG-MAE, we incorporated a multi-branch 1D-CNN architecture. This architecture utilizes filters of varying sizes (1×3, 1×5, and 1×7) to capture multi-scale temporal patterns present in the PSG signals. These extracted features are then concatenated, enabling the network to integrate complementary information from different receptive fields. Subsequently, we apply a dimensionality-reduction step using a 1×1 filter, followed by global pooling, to further distill the feature representation while retaining the most informative components relevant to the task. The final output is then passed through an MLP layer for discrimination, generating task-specific results. A crucial aspect of this approach is the involvement of the pre-trained PSG-MAE encoder during the training phase of the downstream network. By integrating the encoder into the backpropagation process, the network can dynamically fine-tune its feature extraction capabilities. This enables the model to learn features that are specifically tailored to the downstream task. Rigorous validation of the effectiveness of the PSG-MAE encoder is conducted through following experiments on sleep staging and OSA detection tasks.

We employ cross-entropy loss to optimize the models for both downstream tasks. In the sleep staging task, multi-class cross-entropy loss \eqref{cls} is used, while binary cross-entropy loss \eqref{bcls} is applied in the OSA detection task, they are defined as
\begin{equation}
L_{cls} = - \frac{1}{N} \sum_{i=1}^{N} \sum_{j=1}^{C} y_{ij} \log(p_{ij}),
\label{cls}
\end{equation}
\begin{equation}
L_{bcls} = - \frac{1}{N} \sum_{i=1}^{N} \left[ y_i \log(p_i) + (1 - y_i) \log(1 - p_i) \right],
\label{bcls}
\end{equation}
where \( y \) is the true label of sample, \(p\) is the sample's predicted probability of each class. Given the class imbalance in both sleep staging and OSA detection tasks (e.g., sleep apnea events typically constitute a small proportion of the total sleep duration), we apply class weights to the cross-entropy loss function, with higher weights assigned to underrepresented classes to mitigate the impact of class imbalance during training.

\begin{figure*}[!h]
\centering
\includegraphics[width=\textwidth]{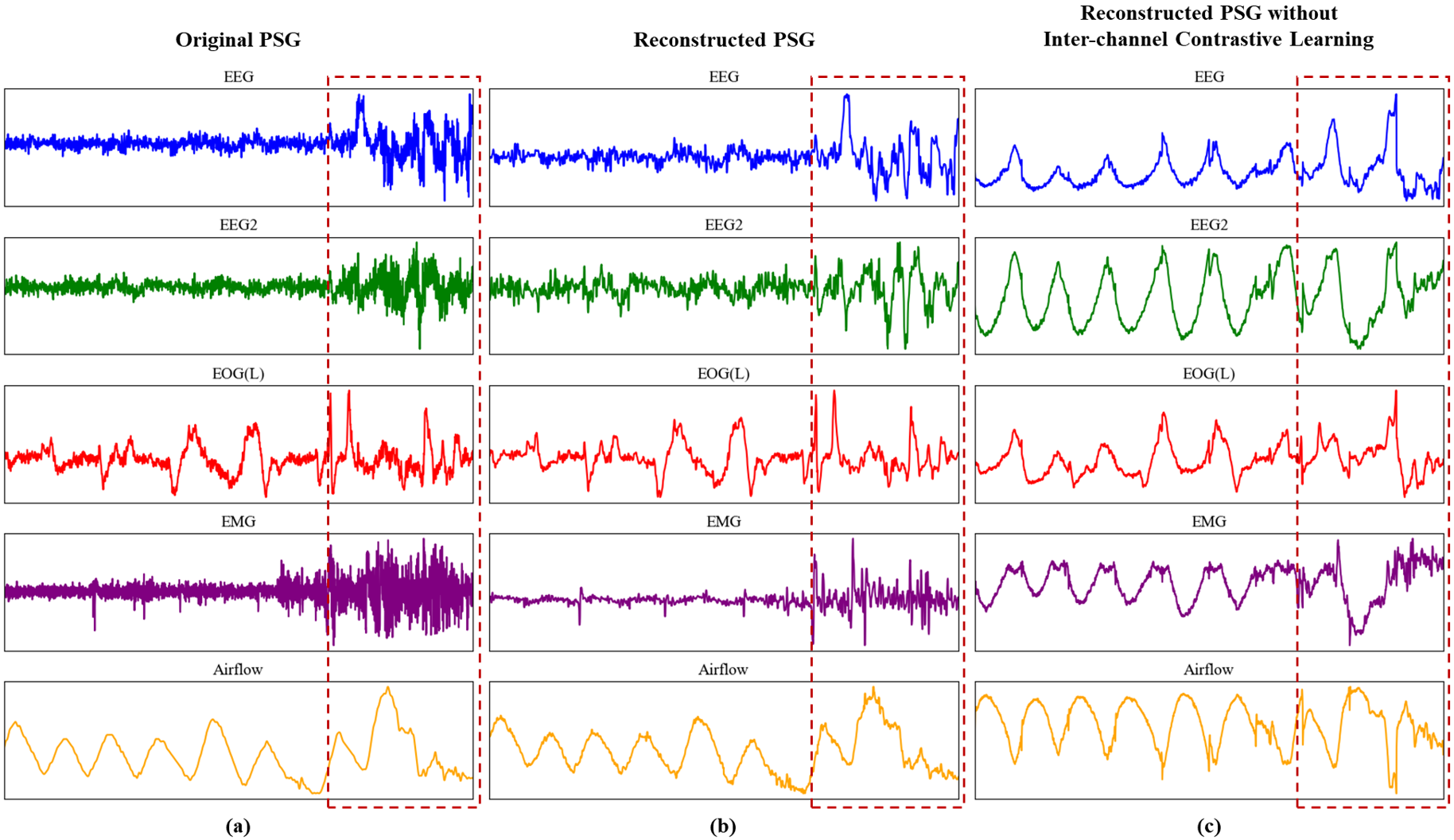}
\caption{The signal reconstruction results of PSG-MAE pre-training, show that the channel-level signal reconstruction loss and ICCL enable the framework to learn fine-grained temporal information within PSG channels as well as interaction information between channels. In contrast, without ICCL, it becomes difficult to disentangle individual channel information from the fused multichannel features.}
\label{fig4}
\end{figure*}

\section{Experiments}

The objective of this study is to achieve robust extraction of both single-channel temporal features and multichannel fusion features from PSG data by proposing the unsupervised learning approach, PSG-MAE. The experiments are designed with two main goals. First, to validate the effectiveness of the PSG-MAE, which uses complementary-masking and ICCL strategies to capture and reconstruct multichannel PSG signal information. Second, to assess the performance of the pre-trained encoder on two downstream sleep event monitoring tasks, sleep staging, and OSA detection, thereby evaluating its feasibility, applicability, and discriminative performance.

\subsection{Dataset}

To ensure sufficient diversity in the PSG data during the pre-training phase of the PSG-MAE and enhance the encoder's robustness and generalization ability, this study uses three different datasets. These datasets not only help in improving the encoder’s performance but also allow for the evaluation of its effectiveness in downstream sleep event monitoring tasks:

The \textbf{Sleep Heart Health Study (SHHS)}\cite{zhang2018national},\cite{10.1093/sleep/20.12.1077} is a multicenter epidemiological research resource aimed at assessing the impact of sleep-disordered breathing on cardiovascular health and other health outcomes. The dataset originates from a study led by the National Heart, Lung, and Blood Institute (NHLBI), with participants from various communities across the United States. It includes approximately 6,000 adults, primarily aged 50 and older, with data collection beginning in 1995 and ongoing long-term follow-up. The dataset covers multiple physiological signal channels with a sampling frequency of 100Hz.

The \textbf{PSG-audio} dataset\cite{korompili2021psg} is sourced from the Sismanoglio – Amalia Fleming General Hospital in Athens, Greece, and was collected and annotated by the hospital's medical team. The dataset contains 212 synchronized PSG recordings, which also include audio recordings of breathing sounds from both tracheal and ambient microphones, for the analysis of apneic events and the development of home-based apnea detection techniques. The sampling frequency of the EEG signals is 200Hz. 

The \textbf{Clinical-PSG} dataset was collected by the Department of Neurology at Guangzhou First People’s Hospital in Guangdong, China, within a clinical laboratory setting. This  It contains PSG data recorded from 371 subjects during their nighttime sleep from 2021 to 2022. The EEG channels have a sampling frequency of 200 Hz, and sleep events are annotated in 30-second sleep epochs by professional sleep medicine physicians. These annotations include sleep stages, apneas, PLMD, and periodic limb movements while awake (PLMA). This dataset is intended for the diagnosis and research of sleep disorders. The construction and utilization processes of this dataset involved no collection of privacy information from subjects and were approved by the Guangzhou First People's Hospital Ethics Committee (Approval No. K-2025-067-01).

\subsection{Experimental Setup}

During the pre-training phase, the synthetic dataset comprises 2,200 nights of PSG data from the SHHS, PSG-audio, and Clinical-PSG datasets. Combining these datasets ensures that the encoder learns the distribution of physiological signals across different datasets and subjects, enhancing the robustness of feature extraction. To maintain consistency in channel selection across the pre-training data, five common channels are chosen from the datasets: right central brain activity, left central brain activity, left eye movement, jaw electromyography, and pressure-based airflow signal. The channel names selected and the amount of data used in the pre-training phase from each dataset are shown in Table \ref{table 1}. The PSG data undergoes EEG artifact removal processing to reduce interference from non-brain signals, thus enhancing the quality and reliability of the data. The sampling frequency of the PSG data is set to 100Hz, therefore the shape of the processed sleep epochs, \((channel\;number,\;data\;length)\) is standardized to \((5, 3000)\). To mitigate individual differences in the PSG data, we apply the Z-Score normalization method to each of the five channels in the PSG signals, the formula is

\begin{equation}
x' = \frac{x - \mu}{\sigma}.
\end{equation}
This approach involves calculating the median \((\mu\)) and standard deviation \((\sigma\)) for each channel, and then using these values to normalize the data accordingly. The training data is randomly shuffled, with 80\% allocated for training, 10\% used for validation, and the remaining 10\% used for testing.

\begin{table}[h]
\caption{Channel Chosen from PSG datasets}
\label{table 1}
\centering
\normalsize
\renewcommand{\arraystretch}{1.2}
\begin{tabular}{|p{1.7cm}|p{1.3cm}|>{\raggedright\arraybackslash}p{4cm}|}
\hline
\textbf{Datasets} & \textbf{Data Number} &\textbf{PSG Channels} \\
\hline
SHHS & 1900 & `EEG (C4-A1)', `EEG2 (C3-A2)', `EOG (L)', `EMG' and `AIRFLOW'\\
\hline
PSG-audio & 200 & `EEG C3-A2', `EEG C4-A1', `EOG LOC-A2', `EMG Chin' and `Flow Patient (Pressure cannula)'\\
\hline
PSG-clinical & 100 & `EEG C3-REF', `EEG C4-REF', `EOG LOC', `EMG Chin' and `Airflow'\\
\hline
\end{tabular}
\end{table}

In subsequent downstream task experiments, the SHHS dataset is utilized for the sleep staging task, where 600 nights of data, not included in the pre-training phase, are randomly selected. In contrast, the remaining 171 nights from the PSG-clinical dataset, with detailed annotations, are employed for OSA detection. The data used in both tasks retains the same channel selection as in the pre-training phase to maintain consistency across tasks. Given the considerable subject-level variability inherent in PSG signals, both downstream tasks employ a subject-wise 5-fold cross-validation strategy. Specifically, the PSG data are randomly partitioned at the subject level into five equal subsets. In each fold, 80\% of the data from four subsets is used for training, 20\% for validation, and the remaining subset is reserved for testing. The final experimental results are obtained by averaging the outcomes across the five folds.

\begin{figure*}[!h]
\centering
\includegraphics[width=0.7\textwidth]{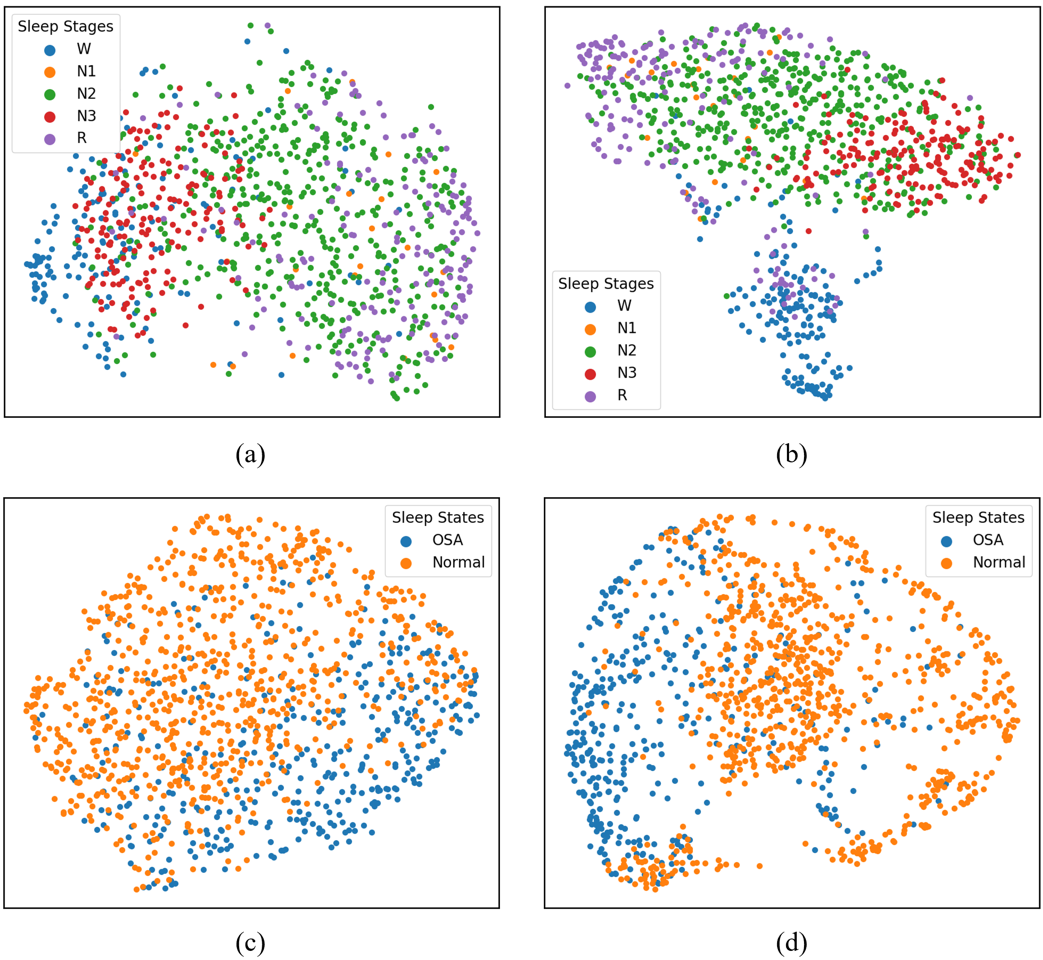}
\caption{UMAP visualization of PSG-MAE extracted features before and after downstream task training: (a) and (b) for sleep staging, (c) and (d) for OSA detection.}
\label{fig 5}
\end{figure*}

\subsection{Evaluation Metrics}

To rigorously evaluate the effectiveness of the upstream pre-training phase of PSG-MAE and its performance in downstream sleep events monitoring tasks, this study firstly employs the mean squared error (MSE) to quantify the discrepancy between the reconstructed and original signals, thereby assessing the accuracy of the signal reconstruction. The MSE is written as
\begin{equation}
MSE = \frac{1}{n} \sum_{i=1}^{n}(y_{\text{pred}} - y_{\text{true}})^2.
\end{equation}

Subsequently, in the sleep staging task, once the true positives (TP), false positives (FP), true negatives (TN), and false negatives (FN) are computed for each class, accuracy (ACC), precision, recall and macro F1-score (MF1) are calculated as

\begin{equation}
ACC = \frac{TP + TN}{TP + TN + FP + FN},
\end{equation}

\begin{equation}
Precision = \frac{TP}{TP + FP},
\end{equation}

\begin{equation}
Recall = \frac{TP}{TP + FN},
\end{equation}

\begin{equation}
F1\text{-}score = \frac{2 \times \text{Precision} \times \text{Recall}}{\text{Precision} + \text{Recall}}.
\end{equation}

Due to the infrequent and brief nature of OSA events, the positive class is underrepresented in the OSA detection task. As a result, accuracy alone may not provide a reliable evaluation. Therefore, MF1 is used alongside accuracy, as this metric offers a more balanced assessment of the model's performance and helps mitigate the bias introduced by the dominant negative samples.

\section{EXPERIMENTAL RESULTS \& DISCUSSION}

\subsection{Results of PSG-MAE Pre-training Process }

\begin{table*}[h]
\caption{The MSE value of PSG channels between the original signal and the reconstructed signal of PSG-MAE.}
\label{table 2}
\centering
\normalsize
\renewcommand{\arraystretch}{1.2}
\begin{tabular}{|p{3cm}|p{2cm}|p{2cm}|p{2cm}|p{2cm}|p{2cm}|}
\hline
\textbf{PSG Channels} & EEG & EEG2 & EOG (L) & EMG & Airflow \\
\hline
\textbf{MSE without ICCL}  & \(2.7 \times 10^{-2}\) & \(2 \times 10^{-2}\) & \(8.9 \times 10^{-2}\) & \(1.5 \times 10^{-2}\) & \(1.69\)\\
\hline
\textbf{MSE with ICCL} & \(\boldsymbol{8 \times 10^{-6}}\) & \(\boldsymbol{7 \times 10^{-6}}\)& \(\boldsymbol{3.6 \times 10^{-5}}\) & \(\boldsymbol{6 \times 10^{-6}}\) & \(\boldsymbol{7.66 \times 10^{-2}}\)\\
\hline
\end{tabular}
\end{table*}

The Fig \ref{fig4}. shows the original PSG signal input and the reconstructed PSG signal output from PSG-MAE during the validation of pre-training phase. A comparison between (a) and (b) shows that the reconstructed signal accurately replicates the trend of signal variations while preserving the temporal details of the original signal. Furthermore, in the highlighted dashing boxes, fluctuations of multiple signal channels in the original signal are aligned with the time scale of the reconstructed signal, indicating that PSG-MAE is capable of capturing underlying relationships among different channels. Through ICCL, PSG-MAE retains the original signal's trend while demonstrating a certain degree of noise suppression, which indicates the interaction content learned by the encoder can help repair corrupted signal channels. The comparison between (2) and (3) underscores the critical role of ICCL in preserving single-channel information within multi-channel fused features. Without ICCL, PSG-MAE struggles to accurately reconstruct signals due to the loss of channel-specific details. This highlights ICCL's ability to enhance the fusion of multi-channel information during the pre-training phase while minimizing information loss from individual channels. Table \ref{table 2} presents the MSE of the selected 5 channels from the reconstructed and original signals, quantitatively reflecting the accuracy of PSG-MAE in signal reconstruction. EEG and EOG signal reconstruction demonstrate good performance, with effective preservation of waveform details and favorable MSE values. Similarly, the PSG-MAE model without ICCL cannot achieve the same level of reconstruction performance. Despite the high noise levels in the original EMG signals, the PSG-MAE model is still able to recover the main trends of the signal. In contrast, airflow signals show greater variability due to the significant time gap between the datasets and variations in data acquisition conditions, which complicate the reconstruction process. The reconstruction performance of some airflow signals is suboptimal, resulting in relatively higher average MSE values.

\begin{table*}[h]
\caption{Methods and Performance Metrics of Sleep Staging Task}
\label{table 3}
\centering
\normalsize
\renewcommand{\arraystretch}{1.5}
\begin{tabular}{>{\centering\arraybackslash}m{4cm}>{\centering\arraybackslash}m{1.3cm}>{\centering\arraybackslash}m{2.4cm}>{\centering\arraybackslash}m{1.5cm}>{\centering\arraybackslash}m{1.5cm}>{\centering\arraybackslash}m{1.5cm}>{\centering\arraybackslash}m{1.5cm}>{\centering\arraybackslash}m{1.5cm}}
\noalign{\hrule height 1.5pt}
\textbf{Methods} & \textbf{Accuracy} & \textbf{Macro F1-score} & \textbf{ACC. W} & \textbf{ACC. N1} & \textbf{ACC. N2} & \textbf{ACC. N3} & \textbf{ACC. R}\\
\hline
SleepEEGNet\cite{mousavi2019sleepeegnet}    & 73.9\% & 68.4\% & 81.3\% & 34.4\% & 73.4\% & 75.9\% & 77.0\%\\
DeepSleepNet\cite{supratak2017deepsleepnet}   & 81.0\% & 73.9\% & 85.4\% & \textbf{40.5\%} & 82.5\% & 79.3\% & 81.9\%\\
MultitaskCNN\cite{8502139}   & 81.4\% & 71.2\% & 82.2\% & 25.7\% & 85.5\% & 83.3\% & 81.1\%\\
AttnSleep\cite{9417097}     & 82.3\% & 74.1\% & 85.0\% & 34.2\% & 85.7\% & 83.5\% & 82.3\%\\
CausalAttenNet\cite{10663435} & 83.3\% & 73.1\% & 85.4\% & 27.6\% & \textbf{84.8\%} & 84.0\% & \textbf{82.9\%}\\
\textbf{Ours}        & \textbf{83.7\%} & \textbf{74.7\%} & \textbf{94.8\%} & 33.2\% & 84.5\% & \textbf{85.3}\% & 79.8\%\\
\noalign{\hrule height 1.5pt}
\end{tabular}
\end{table*}

\subsection{Downstream Sleep Event Monitoring Results}

The Fig \ref{fig 5}. shows the uniform manifold approximation and projection (UMAP) visualization of features extracted by the pre-trained encoder. Each point represents the reduced feature of one sleep epoch. The feature distribution before and after the training of the sleep staging task is shown in (a) and (b) respectively. A comparison reveals that, before the downstream task training, the features extracted by the pre-trained encoder exhibit only subtle clustering. After the downstream task training, features corresponding to different sleep stages exhibit more distinct clustering, indicating that the encoder has been fine-tuned to better align with the feature extraction requirements of the downstream task. Table \ref{table 3} provides the sleep staging performance of the fine-tuned encoder, coupled with the downstream classification head. Compared to other approaches, the proposed PSG-MAE framework demonstrates superior performance in terms of prediction accuracy and MF1 score, showing enhanced discrimination, particularly for the W (waking) and N3 (deep sleep) stages.

\begin{table}[h!]
\caption{Methods and Performance Metrics for OSA detection}
\label{table 4}
\centering
\normalsize
\renewcommand{\arraystretch}{1.5}
\begin{tabular}{>{\centering\arraybackslash}m{2cm}>{\centering\arraybackslash}m{2cm}>{\centering\arraybackslash}m{2cm}}
\noalign{\hrule height 1.5pt}
\textbf{Methods} & \textbf{Accuracy} & \textbf{Macro F1-score} \\
\hline
RF & 83.7\% & 56.1\% \\
SVM & 87.1\% & 46.5\% \\
CNN & 87.6\% & 46.7\% \\
\textbf{Ours} & \textbf{90.45\%} & \textbf{67.33\%} \\
\noalign{\hrule height 1.5pt}
\end{tabular}
\end{table}

The UMAP visualization in (c) and (d) of OSA detection training on the PSG-clinical dataset shows that features of OSA epochs and normal epochs are intermixed in the feature space before training and then a more distinct separation between the two classes emerges after the downstream fine-tuning of the encoder, indicating that the fine-tuned encoder is capable of distinguishing between OSA and normal sleep states. Table \ref{table 4} presents the performance of the fine-tuned encoder combined with the downstream network for OSA detection. The model is compared to the machine learning methods of random forest (RF) and support vector machine (SVM) and a normal 1D-CNN network. These comparative models also take the raw PSG signals as input and the proposed approach shows a clear advantage in diagnostic accuracy and the MF1 score reaches its optimal value, demonstrating that the proposed network can effectively detect OSA even in the case of imbalanced label distribution.

\subsection{Discussion}

Experimental results validate the effectiveness of the PSG-MAE pre-training process and its feasibility for adaptation to downstream sleep event monitoring tasks. Specifically, by adding a pair of complementary masks to multiple channels of PSG segments from the same time subsegment, the channel-level signal reconstruction loss ensures that the encoder can learn to extract temporal features from multiple channels in the signal reconstruction process. Additionally, the complementary channels engage in ICCL with channels from different time segments, enabling the encoder to capture inter-channel interaction information at the same time point. Experimental results demonstrate that, for noisy channels, ICCL helps clarify the trend of reconstructed signal variations, and ensure simultaneously reconstructed signals involving multiple channels can preserve channel-related information and align with time steps. During the training of downstream sleep event monitoring tasks, the pre-trained encoder, when combined with the downstream feature decomposing network and fine-tuned, shows strong performance in both sleep staging and OSA detection. This indicates that PSG-MAE is not only suitable for the current tasks but also has the potential to adapt to a variety of downstream tasks, thereby facilitating the development of a multi-dimensional system for comprehensive sleep assessment.

\section{CONCLUSION}

We propose an MAE-based pre-training framework named PSG-MAE, which leverages unlabeled PSG data through self-supervised learning to enhance the feature extraction capability of automated sleep event monitoring networks. In the pre-training phase, the channel-level signal reconstruction loss ensures the extraction of fine-grained time-series features, while ICCL emphasizes channel interaction information. The training process adopts PSG data from different datasets with the same channel configuration to improve adaptability to diverse data distributions. By integrating with downstream task networks, the pre-trained encoder can be fine-tuned to perform multitask sleep event monitoring. Compared to traditional single-task sleep monitoring networks, PSG-MAE demonstrates greater versatility. Experimental results show that the PSG-MAE pre-training framework can learn stable PSG features and achieve remarkable performance in both sleep staging and OSA detection tasks. The current limitation of PSG-MAE framework lies in the suboptimal performance of fine-tuned sleep staging network in recognizing N1 and R sleep stages. This issue may stem from the relatively small proportion of N1 and R stages in the training dataset and the lack of targeted strategies for data augmentation. In the future, efforts will directed toward refining the downstream network and training process to improve sleep staging accuracy. Additionally, the framework will be expanded to support a broader range of automated sleep event monitoring tasks. The ultimate objective is to develop a multidimensional sleep monitoring system capable of delivering a more comprehensive analysis of PSG signals from a single input segment.

\section*{REFERENCES}

\bibliographystyle{IEEEtran}
\bibliography{ref.bib}

\end{document}